\documentclass[10pt,twocolumn,letterpaper]{article}

\usepackage[algorithms]{wacv}

\newcommand{\rev}[1]{#1}

\usepackage{graphicx}
\usepackage{amsmath, amssymb, mathtools}
\usepackage{booktabs}
\usepackage{multirow}
\usepackage{array}
\usepackage{colortbl, xcolor}
\usepackage{siunitx}

\definecolor{advitgray}{gray}{0.92}
\newcommand{\best}[1]{\textbf{#1}}
\newcommand{\secbest}[1]{\underline{#1}}
\newcommand{\gph}[1]{\textcolor{gray}{#1}}
\newcommand{\codeavail}{%
  \hyperlink{codeavail}{\textsuperscript{\normalfont*}}%
}

\definecolor{wacvblue}{rgb}{0.21,0.49,0.74}
\usepackage[pagebackref,breaklinks,colorlinks,allcolors=wacvblue]{hyperref}

\def\wacvPaperID{3483} 
\def\confName{WACV}
\def\confYear{2027}

\title{A Case for Axis-Decoupled, Channel-Primary Vision Transformers\\
without Auxiliary Modules}

\author{
Aon Safdar\\
University College Dublin, Ireland\\
{\tt\small aon.safdar@ucdconnect.ie}
\and
Mohamed Saadeldin\\
University College Dublin, Ireland\\
{\tt\small mohamed.saadeldin@ucd.ie}
}

\begin{document}
\AddToShipoutPicture*{%
  \AtPageLowerLeft{%
    \hspace{\dimexpr 1in+\oddsidemargin\relax}%
    \raisebox{0.58in}{%
      \begin{minipage}{\columnwidth}
        \raggedright\footnotesize
        \rule{0.4\columnwidth}{0.4pt}\\[2pt]
        \hypertarget{codeavail}{\textsuperscript{*}\,Code and model weights: \url{https://github.com/aonsafdar/advit}.}%
      \end{minipage}%
    }%
  }%
}
\maketitle
\raggedbottom

\begin{abstract}
\rev{Self-attention in Vision Transformers (ViTs) mixes information along the spatial-token axis, an operator quadratic in token count with weak locality priors, leaving high-resolution stages expensive and small-data training fragile. The dominant response keeps spatial attention as the primary mixer and adds machinery around it: hierarchical tokenization, convolutional stems and $Q/K/V$, or secondary channel branches. Even channel-aware models typically change only the axis while inheriting that spatial-first pipeline. We ask how far a deliberately simple alternative can go: make content-adaptive mixing \emph{channel-primary} and co-design the stage around it rather than grafting channel attention onto a spatial backbone. An axis-decoupling lens views a stage through four design dimensions and assigns its two jobs to purpose-matched operators: depthwise convolutions extract spatial structure, while single-head self-attention over channel tokens mixes information adaptively, global context at linear spatial cost, without auxiliary modules. Trained from scratch, ADViT\codeavail{} leads all three from-scratch benchmarks, gaining $+5.9$ on CIFAR-100 and $+4.6$ on Tiny-ImageNet over larger spatial-attention ViTs; on ImageNet-1K it comes within $0.8$ points of the strongest channel-aware ViTs at $40\%$ fewer parameters, with the best accuracy-efficiency trade-off on ADE20K. We offer the lens and this evidence as a case for channel-primary mixing as first-class backbone design.}
\end{abstract}

\section{Introduction}
\label{sec:intro}

\begin{figure}[t]
    \centering
    \includegraphics[width=\linewidth]{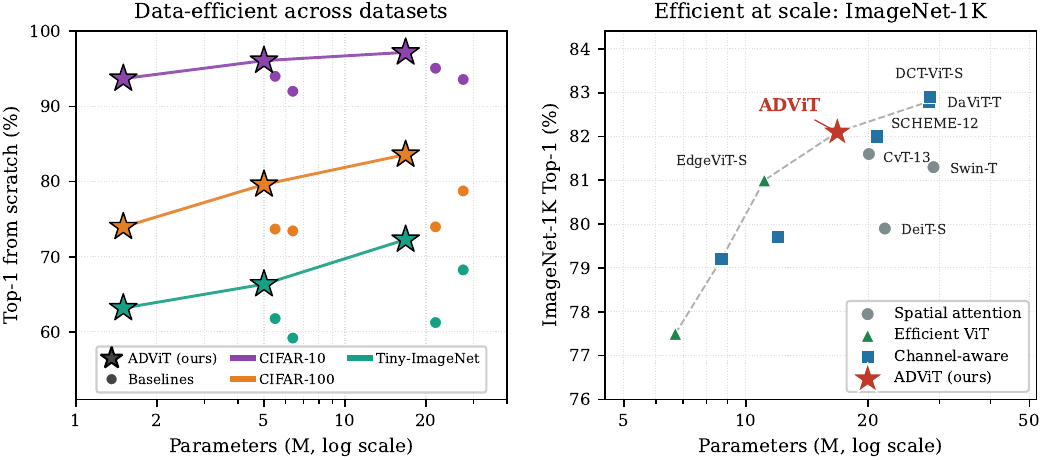}
    \caption{\textbf{ADViT is parameter- and data-efficient.}
    \emph{Left:} trained from scratch on CIFAR-10, CIFAR-100, and Tiny-ImageNet, the ADViT family (stars) sits above and to the left of spatial-attention ViTs (dots) at matched or smaller parameter budgets, with the largest margins on CIFAR-100 and Tiny-ImageNet.
    \emph{Right:} on ImageNet-1K, ADViT-S matches the strongest channel-aware models within $0.8$ points at $\sim$$40\%$ fewer parameters, while remaining competitive with spatial-attention and efficient ViTs at comparable budgets. All numbers are from Tables~\ref{tab:fromscratch} and~\ref{tab:imagenet}.}
    \label{fig:teaser}
\end{figure}

Self-attention in Vision Transformers (ViTs)~\cite{newdosovitskiy2021an,touvron_training_2021} mixes information along the \emph{spatial-token} axis, so every patch attends to every other patch, producing content-adaptive spatial mixing. This single operator is asked to do two conceptually distinct jobs at once, \emph{extracting} local spatial structure and \emph{mixing} information across the representation, and it pays for this twice over. Its cost grows quadratically with the token count, so high-resolution stages are expensive~\cite{liu_swin_nodate,avidan_maxvit_2022}, and because early single-shot patchification discards fine spatial structure, the resulting models are data hungry and brittle when trained from scratch on modest datasets.

Efforts to address these symptoms almost always retain spatial attention as the primary mixer. Hierarchical tokenization spreads spatial reduction across stages to preserve structure~\cite{liu_swin_nodate,wu_cvt_2021,wang_pyramid_2021}; hybrids inject locality through convolutional stems, local modules, or convolutional feed-forward layers~\cite{dai_coatnet_2021,peng_conformer_2021,haruna_exploring_2025}, or push it into the attention inputs through convolutional and depthwise $Q/K/V$ projections~\cite{wu_cvt_2021,cao_visual_2025}. A smaller line revisits the mixing axis itself. Because attention is a content-adaptive mixing operator, it can mix \emph{channels} rather than spatial tokens~\cite{ali_xcit_2021,ding_davit_2022,guan_dynamic_2025,zheng_lightweight_2023,sridhar_scheme_2024}. Yet these methods graft channel attention onto a pipeline still built for spatial mixing, so the tokenization, projections, and locality modules are inherited from spatial-attention backbones, and the two axes are interleaved, fused, or gated. Channel attention remains a secondary branch rather than the organizing principle of the network.

We instead ask how far a deliberately simple alternative can go, making content-adaptive mixing \emph{channel-primary} and \rev{decoupling} the two axes, co-designing the stage around channel attention instead of interleaving, fusing, or gating it with spatial attention. The two jobs that spatial attention conflates are separated and assigned to purpose-matched operators. Spatial structure is extracted by cheap depthwise convolutions that keep the spatial grid intact throughout depth, and information is mixed by self-attention over \emph{channel tokens}. A channel token is a feature-map descriptor that summarizes the entire spatial grid, so attention over channel tokens learns input-dependent inter-feature correlations across the whole image, for instance that ``fur texture'' should up-weight ``animal shape'' and suppress ``background foliage,'' at a cost of $O(HW\,C^2)$, linear in the token count, rather than the $O((HW)^2C)$ of spatial attention.

\begin{figure}[t]
    \centering
    \includegraphics[width=\linewidth]{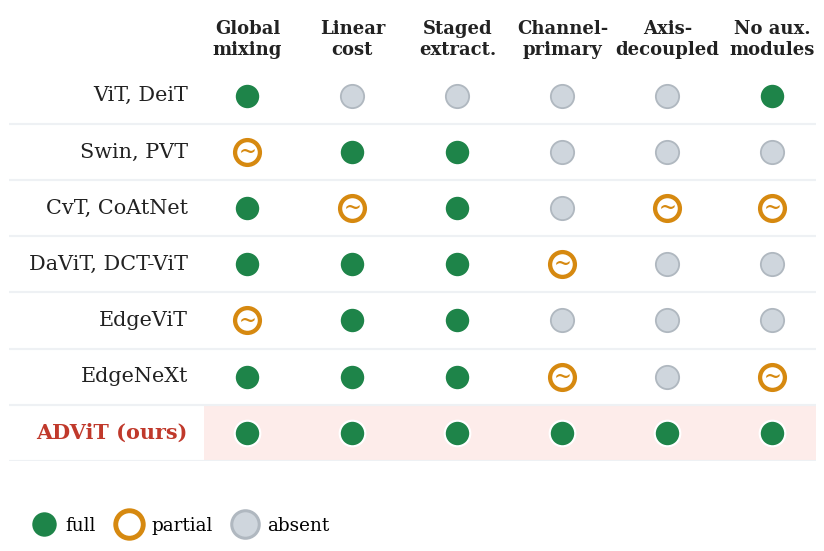}
    \caption{\textbf{Capabilities of backbone families.}
    Each row is a representative family; each column a design property for efficient, data-limited recognition. Prior designs satisfy some properties at the expense of others. \rev{Among the families shown, ADViT attains all six. The columns are design properties, not a ranking: a backbone need not hold all six to be effective, and the figure locates designs in this space rather than scoring them.} ``Partial'' marks a property held only locally, only as a secondary branch, \rev{or only in part}.}
    \label{fig:capability}
\end{figure}

The contribution is not channel attention, which is well known, but the finding that making it primary pays off only when the whole stage is rebuilt around it. We analyze a stage through four design dimensions, (i)~tokenization and downsampling, (ii)~non-attention projections (locality modules and the MLP), (iii)~attention projections ($Q/K/V$ formation), and (iv)~the mixing axis. Prior channel-aware models change only the axis (iv) while inheriting (i)--(iii) from spatial baselines (Table~\ref{tab:lit_mapping}). Co-designing all four around channel mixing has two effects. First, each operator is matched to its role, with depthwise extraction that adds no channel mixing, depthwise-only $Q/K/V$ that add spatial context without prematurely mixing the very channels attention will correlate, and pointwise MLPs for channel capacity. Second, the specialized machinery that dual-axis designs require becomes unnecessary, because channel attention is already global at linear spatial cost, so ADViT uses no windowing~\cite{liu_swin_2021}, no local-patch-interaction module~\cite{ali_xcit_2021}, no alternating dual-axis stages~\cite{ding_davit_2022}, and no dynamic gating~\cite{guan_dynamic_2025}. ADViT combines global mixing, linear spatial cost, staged spatial extraction, channel-primary attention, and freedom from auxiliary modules in a single design (Fig.~\ref{fig:capability}).

This design is most advantageous where ViTs are weakest, namely when parameters, compute, or data are limited (Fig.~\ref{fig:teaser}). Trained from scratch, \rev{ADViT-XS improves over ViT-Ti by $4.6$ points on Tiny-ImageNet and $5.9$ on CIFAR-100 at $9\%$ fewer parameters, and ADViT-S improves over CvT-13 by $1.6$ points on Tiny-ImageNet at $16\%$ fewer parameters}; on ImageNet-1K, ADViT-S is competitive with spatial-attention models of similar size and matches DaViT-T and DCT-ViT-S to within $0.8$ points at $40\%$ fewer parameters. The same backbone gives the best accuracy-efficiency trade-off on ADE20K segmentation and the strongest fine-grained transfer in its size class.

Our contributions are:
\begin{itemize}
\itemsep0.1em
\item An \textbf{axis-decoupling lens}, a four-dimension view of a transformer stage (tokenization, non-attention projections, attention projections, and mixing axis) that clarifies where adaptive mixing should live and makes precise how prior channel-aware models change only the mixing axis while inheriting a spatial-first pipeline. We intend this as a vocabulary the community can reuse to reason about channel-first design.
\item \textbf{\rev{Axis-decoupled, channel-primary mixing}}, realized as \textbf{ADViT}: spatial extraction and channel mixing go to purpose-matched operators, removing the windowing, local-interaction, dual-axis, and gating modules of prior channel-aware ViTs, with a compute model for when this is favorable.
\item Evidence that this simplicity is not paid for in accuracy: across from-scratch recognition, ImageNet-1K, transfer, and dense prediction, the \rev{channel-primary} design is competitive with interleaved and dual-axis channel models and is most advantageous when parameters, compute, or data are limited.
\end{itemize}

\section{Related Work}
\label{sec:related}

\begin{figure}[t]
    \centering
    \includegraphics[width=0.90\linewidth]{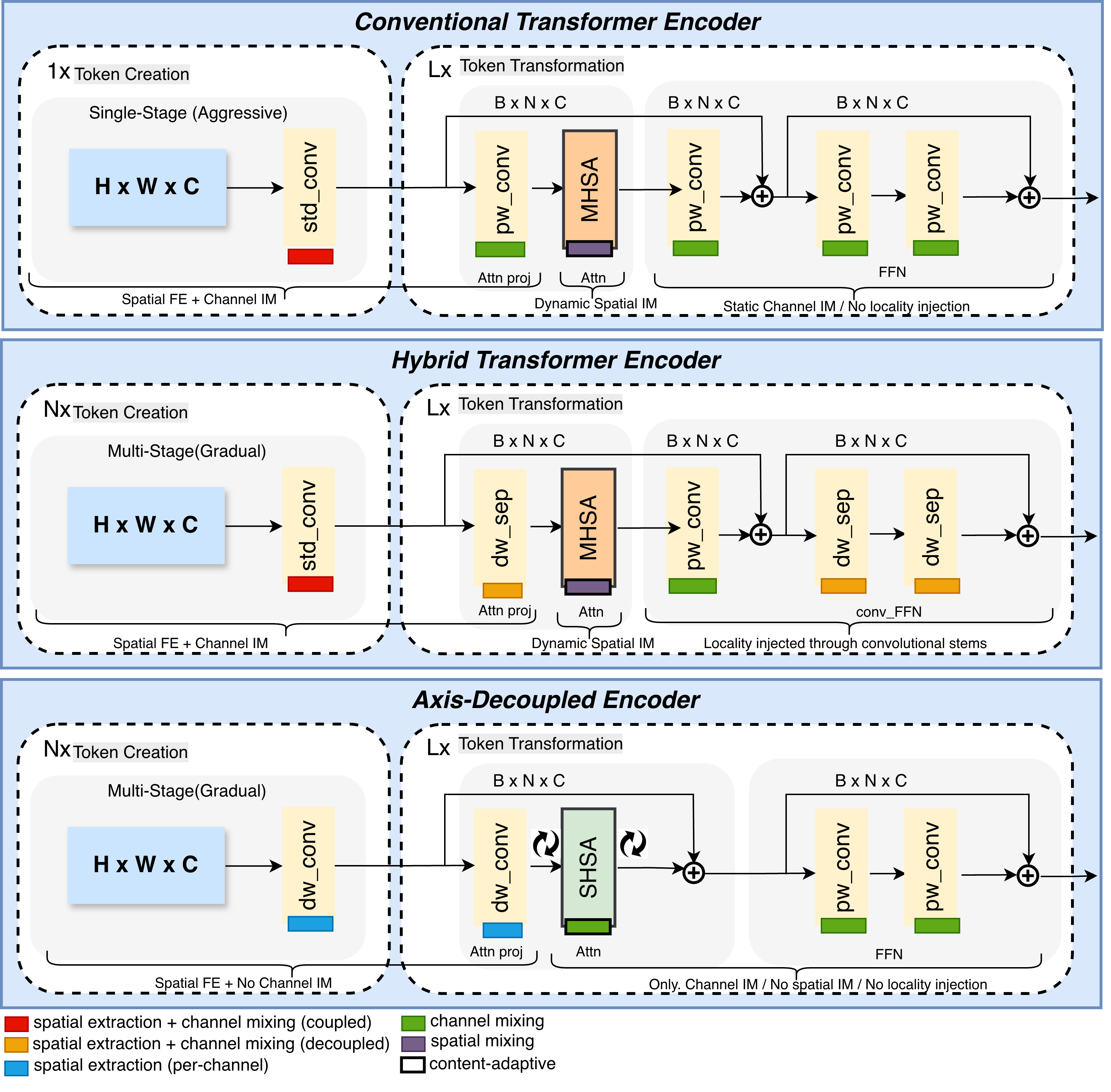}
    \caption{\textbf{Block-level comparison.}
    Top: conventional ViT (spatial attention, pointwise projections).
    Middle: hybrid (spatial attention, convolutional $Q/K/V$).
    Bottom: the ADViT block (channel attention, depthwise-only $Q/K/V$, separate spatial extraction via SRCE).
    The legend distinguishes \emph{extraction} from \emph{mixing} and \emph{static} from \emph{content-adaptive} operators.}
    \label{fig:blocks}
\end{figure}

We organize prior work along the four design dimensions introduced in Section~\ref{sec:intro} and summarized in Table~\ref{tab:lit_mapping}, namely tokenization, non-attention projections, attention projections ($Q/K/V$), and mixing axis. Figure~\ref{fig:blocks} contrasts a conventional ViT block, a hybrid block, and the ADViT block at a glance.

\paragraph{Spatial-attention backbones.}
Early ViTs~\cite{newdosovitskiy2021an,touvron_training_2021} use single-shot patch tokenization with global spatial attention and pointwise MLPs. Hierarchical variants such as Swin~\cite{liu_swin_2021,liu_swin_nodate}, PVT~\cite{wang_pyramid_2021}, and CvT~\cite{wu_cvt_2021} distribute spatial reduction across stages for multi-scale features. A broad class of hybrids injects locality \emph{outside} attention through convolutional stems, local modules, or convolutional feed-forward layers~\cite{dai_coatnet_2021,peng_conformer_2021,han_survey_2023}, while another pushes locality \emph{into} the attention inputs through convolutional or depthwise-separable $Q/K/V$ projections~\cite{wu_cvt_2021,cao_visual_2025}. In all of these the mixing axis stays spatial, and the projection and tokenization choices are tuned for spatial attention.

\paragraph{Efficient and mobile ViTs.}
A parallel effort targets tight compute budgets directly. MobileViT~\cite{mehta_mobilevit_2022} and EdgeViT~\cite{pan_edgevits_2022} interleave lightweight convolutions with sparse attention, and SHViT-style single-head designs~\cite{yun_shvit_2024} reduce head redundancy. These methods improve the accuracy-latency trade-off but retain spatial attention as the core mixer; ADViT instead changes which axis carries the adaptive mixing, and we compare against this family directly (Table~\ref{tab:imagenet}).

\paragraph{Channel-axis and dual-axis attention.}
Because attention is a content-adaptive mixing operator, it can be applied over channel tokens. XCiT~\cite{ali_xcit_2021} introduces cross-covariance attention over channels, supplemented by a depthwise local-patch-interaction (LPI) module. DaViT~\cite{ding_davit_2022} alternates windowed spatial attention with grouped channel attention. SCAT~\cite{zheng_lightweight_2023} couples reduced-resolution spatial attention with channel attention, and DCT-ViT~\cite{guan_dynamic_2025} dynamically gates windowed spatial and channel self-attention. Closest in spirit, SCHEME~\cite{sridhar_scheme_2024} replaces the dense MLP with a block-diagonal channel mixer, and asymmetric factorized bilinear operators~\cite{wu_asymmetric_2025} enrich channel interactions inside the token mixer. \rev{Two works are architecturally closest. EdgeNeXt~\cite{maaz_edgenext_2022} is a ConvNeXt hybrid whose SDTA encoder entangles extraction and mixing in one fused block (channel split, pointwise $Q/K/V$), so it is not axis-decoupled. Restormer~\cite{zamir_restormer_2022} applies transposed $C{\times}C$ attention with depthwise-convolutional $Q/K/V$ for image restoration. ADViT differs on four ablated points: (i)~$Q/K/V$ are \emph{purely depthwise}, whereas Restormer prepends a $1{\times}1$ and EdgeNeXt/XCiT use pointwise projections (Table~\ref{tab:ablation}); (ii)~SRCE and CSA occupy \emph{separate} blocks, whereas SDTA fuses them by channel splitting; (iii)~the mixing axis varies with resolution; (iv)~a single head replaces grouped multi-head attention. Restormer further targets restoration rather than recognition.} These works show that channel mixing is valuable, but in \rev{most} cases it sits in a spatial-first pipeline where the two axes are interleaved, fused, or gated rather than cleanly separated.

\begin{table}[t]
\centering
\caption{\textbf{Design-dimension comparison.}
Architectures viewed through the four design dimensions of Section~\ref{sec:intro}.
\rev{Most prior channel-aware models retain a spatial-first pipeline, inheriting tokenization and $Q/K/V$ and interleaving, fusing, or gating the two axes.}
ADViT redesigns the full pipeline around channel attention as the primary mixer.
PW: pointwise; DW: depthwise; DW-sep: depthwise-separable.}
\label{tab:lit_mapping}
\setlength{\tabcolsep}{3pt}
\renewcommand{\arraystretch}{1.08}
\footnotesize
\resizebox{\linewidth}{!}{%
\begin{tabular}{@{}l l l l l@{}}
\toprule
\textbf{Model} & \textbf{Token.} & \textbf{Non-attn proj.} & \textbf{$Q/K/V$} & \textbf{Mixing axis} \\
\midrule
ViT~\cite{newdosovitskiy2021an}    & Single  & PW MLP            & PW       & Spatial (global) \\
DeiT~\cite{touvron_training_2021}  & Single  & PW MLP            & PW       & Spatial (global) \\
Swin~\cite{liu_swin_2021}          & Staged  & PW MLP            & PW       & Spatial (window) \\
CvT~\cite{wu_cvt_2021}             & Staged  & Conv $+$ PW MLP   & DW-sep   & Spatial \\
CoAtNet~\cite{dai_coatnet_2021}    & Staged  & Conv $+$ PW MLP   & PW       & Spatial \\
XCiT~\cite{ali_xcit_2021}          & Single  & DW LPI $+$ PW MLP & PW       & Channel $+$ local \\
DaViT~\cite{ding_davit_2022}       & Staged  & PW MLP            & PW       & Spatial $/$ channel \\
SCAT~\cite{zheng_lightweight_2023} & Pyramid & Conv $+$ PW MLP   & Standard & Spatial $+$ channel \\
DCT-ViT~\cite{guan_dynamic_2025}   & Pyramid & PW MLP            & Dyn.\ ch.& Spatial $+$ channel \\
SCHEME~\cite{sridhar_scheme_2024}  & Staged  & Block-diag.\ MLP  & PW       & Spatial \\
\rev{EdgeNeXt~\cite{maaz_edgenext_2022}} & \rev{Staged} & \rev{Multi-scale DW $+$ PW MLP} & \rev{PW} & \rev{Fused DW $+$ channel} \\
\rev{Restormer~\cite{zamir_restormer_2022}} & \rev{U-Net} & \rev{Gated DW FFN} & \rev{DW-sep} & \rev{Channel} \\
\midrule
\rowcolor{advitgray}
\textbf{ADViT} & Staged & DW SRCE $+$ PW MLP & DW-only & \textbf{Channel-primary} \\
\bottomrule
\end{tabular}}
\end{table}

\paragraph{Positioning.}
ADViT treats channel attention as the primary representation-learning mechanism rather than a supplement to spatial attention, and redesigns the full pipeline around it (Table~\ref{tab:lit_mapping}): SRCE extracts without channel mixing, depthwise-only $Q/K/V$ avoid mixing before CSA, and spatial attention is used at the last, low-resolution stage, where the lens assigns it as complementary to channel mixing. To our knowledge, no prior backbone co-designs all four dimensions around channel attention as the primary mixer.

\section{Method}
\label{sec:method}

We first motivate axis-decoupled channel-first mixing and derive the operator choices it implies (Section~\ref{subsec:why}), then describe the ADViT architecture (Section~\ref{subsec:arch}).

\subsection{Axis-Decoupled Channel-First Mixing}
\label{subsec:why}

\paragraph{Why decoupling helps.}
Spatial attention is asked to do two jobs with one operator, to \emph{extract} local spatial structure and to \emph{mix} information globally. It is well matched to neither. It carries only a weak locality prior, so the extraction job is data-hungry, and its cost is quadratic in the token count, so the mixing job is expensive at high resolution. Decoupling the two jobs lets each go to an operator suited to it. Extraction is handed to depthwise convolution, which has a strong, parameter-cheap locality prior and is therefore data-efficient; long-range mixing is handed to attention over channel tokens, which is content-adaptive and \rev{image-wide, since each channel token summarizes the whole image so the resulting $C{\times}C$ mixing is applied identically at every location,} yet costs $O(HW\,C^2)$, linear in the spatial size. The benefit is not that channel attention extracts spatial structure (it does not), but that separating the roles frees each operator to do one job well, which is why the design is simultaneously more data-efficient and cheaper at high resolution.

\rev{\noindent\textbf{Axis decoupling, operationally.} We call an operator \emph{spatial} if it has non-trivial receptive field but no cross-channel term, so each output channel depends only on the corresponding input channel (\eg{} depthwise convolution), and \emph{channel} if it is position-wise, applied identically at every location, and therefore extracts no new spatial structure (\eg{} $1{\times}1$ projections, MLPs, and the $C{\times}C$ channel attention of Eq.~\eqref{eq:channel_attn}). A backbone is \emph{axis-decoupled} when no single operator does both, so every block is either spatial or channel, never a fused operator that simultaneously extracts spatial structure and mixes channels. ADViT's backbone satisfies this by construction (depthwise SRCE for extraction, position-wise CSA/MLP for mixing). The same lens treats channel and spatial attention as complementary, and therefore schedules spatial mixing only at the last, low-resolution stage. We describe the backbone as \emph{axis-decoupled} and the full model as \emph{channel-primary} rather than spatial-first. We use three terms precisely. \emph{Channel mixing} is any cross-channel operation, whether a pointwise MLP, a $1{\times}1$ convolution, or channel attention. \emph{Channel attention} (CSA) is our \emph{content-adaptive} channel mixer, the $C{\times}C$ self-attention of Eq.~\eqref{eq:channel_attn}, as distinct from the static mixing of an MLP or the per-channel gating of squeeze-and-excitation. \emph{Channel-primary} representation learning makes this content-adaptive channel mixing the network's principal mixer rather than a secondary branch beside spatial attention.}

\paragraph{Stages as extraction-mixing pairs.}
We view each backbone stage as a mapping
\begin{equation}
V_{\mathrm{out}} = M_{\alpha}\!\bigl(E(V_{\mathrm{in}})\bigr),
\label{eq:projection}
\end{equation}
where $E(\cdot)$ is a \emph{spatial feature-extraction} operator (local aggregation, optionally with downsampling) and $M_{\alpha}(\cdot)$ is an \emph{information-mixing} operator along axis $\alpha\!\in\!\{\text{spatial},\text{channel}\}$. This is an analysis lens. Stages may interleave the two in practice, but the decomposition makes explicit \emph{what is extracted}, \emph{what is mixed}, and \emph{along which axis}. It exposes a degree of freedom prior work leaves implicit, namely that $\alpha$ can vary across stages and that the operators for $E$ and $M_\alpha$ should be matched to their roles.

\paragraph{Channel tokens abstract the whole image.}
Consider a feature volume $X\!\in\!\mathbb{R}^{H\times W\times C}$. In spatial attention, each of the $N{=}HW$ tokens is a $C$-dimensional descriptor of a local patch, and attention discovers relationships \emph{between patches}. If instead we treat the $C$ feature maps as tokens, each one is a descriptor that summarizes the \emph{entire} spatial grid for a single feature dimension. Self-attention over channel tokens computes a $C{\times}C$ correlation matrix,
\begin{equation}
A = \mathrm{softmax}\!\left(\frac{QK^{\!\top}}{\sqrt{d}}\right)\!\in\!\mathbb{R}^{C\times C},
\label{eq:channel_attn}
\end{equation}
which re-weights every channel at every spatial location through a broadcast $Y_{h,w,:}{=}A\,V_{h,w,:}$. This learns input-dependent inter-feature correlations (``when texture $i$ is active, enhance shape $j$'') rather than inter-region routing. Because each channel token already sees the full image, a single attention head suffices to capture global feature interactions without fragmenting them into low-dimensional sub-spaces.

\paragraph{Projections are channel mixers.}
A key observation for decoupled design is that all pointwise ($1{\times}1$) projections, including MLP layers and the linear maps that form $Q/K/V$, are position-wise \emph{channel mixers}, computing $y_{i,j}{=}Wx_{i,j}{+}b$ for $x_{i,j}\!\in\!\mathbb{R}^{C}$ at every $(i,j)$. They mix channels identically at every location and contribute \emph{no} spatial extraction. A backbone built only from pointwise projections and spatial attention therefore performs all of its feature extraction during tokenization, and no new spatial features are extracted afterwards.

\paragraph{Decoupling enables purpose-matched operators.}
In existing dual-axis designs~\cite{ding_davit_2022,zheng_lightweight_2023,guan_dynamic_2025}, spatial and channel attention share a projection and tokenization pipeline designed for spatial mixing. Decoupling the axes lets us match each operator to its role, guided by the compute profile in Table~\ref{tab:compute_profile}:
\begin{itemize}
\itemsep0.1em
\item \textbf{Spatial extraction (SRCE):} pure spatial aggregation with no channel mixing calls for a depthwise $k{\times}k$ convolution, the only operator with spatial extent but no cross-channel term.
\item \textbf{$Q/K/V$ formation (CSA):} since channel attention will mix channels, a pointwise projection beforehand would add \emph{redundant} channel mixing ($HW\,C^2$ extra FLOPs) and corrupt the raw feature-map descriptors. Depthwise-only projections inject spatial context at minimal cost with no premature channel interaction.
\item \textbf{Non-attention projection (MLP):} the post-attention MLP has a purely channel-mixing role, so pointwise is exactly right.
\item \textbf{Stage schedule:} applying the $C{\times}C$ matrix at every location costs $O(HW\,C^2)$, the same as a pointwise projection, and carries \emph{no} $(HW)^2$ term; spatial attention costs $O((HW)^2C{+}HW\,C^2)$. The saving is therefore the quadratic spatial term, whose size relative to channel mixing is $HW/C$: $49{\times}$ at stage~1 ($HW{=}3136$, $C{=}64$) but only $0.5{\times}$ at stage~3 ($HW{=}196$, $C{=}384$). Spatial attention is thus prohibitive at high resolution and affordable at the last stage, where we schedule it.
\end{itemize}

\begin{table}[t]
\centering
\caption{\textbf{Compute profile of convolutional operators.}
$V_{\text{in}}\!\in\!\mathbb{R}^{H\times W\times C_{\text{in}}}$, $V_{\text{out}}\!\in\!\mathbb{R}^{H\times W\times C_{\text{out}}}$.
\checkmark/-- indicate whether the operator performs spatial extraction or channel mixing.
For $C_{\text{in}}\!\approx\!C_{\text{out}}$ and $k{>}1$ the cost ordering is standard $>$ DW-sep $>$ pointwise $>$ depthwise-only.}
\label{tab:compute_profile}
\scriptsize
\setlength{\tabcolsep}{3pt}
\renewcommand{\arraystretch}{1.12}
\resizebox{\linewidth}{!}{%
\begin{tabular}{@{}lcccc@{}}
\toprule
\textbf{Operator} & \textbf{Spatial} & \textbf{Channel} & \textbf{Params} & \textbf{FLOPs} \\
\midrule
Standard $k{\times}k$  & \checkmark & \checkmark    & $k^{2}C_{\text{in}}C_{\text{out}}$               & $HW\,k^{2}C_{\text{in}}C_{\text{out}}$ \\
Depthwise-sep          & \checkmark & \checkmark & $k^{2}C_{\text{in}}{+}C_{\text{in}}C_{\text{out}}$  & $HW(k^{2}C_{\text{in}}{+}C_{\text{in}}C_{\text{out}})$ \\
Pointwise $1{\times}1$ & --        & \checkmark    & $C_{\text{in}}C_{\text{out}}$                     & $HW\,C_{\text{in}}C_{\text{out}}$ \\
Depthwise $k{\times}k$ & \checkmark & --           & $k^{2}C_{\text{in}}$                              & $HW\,k^{2}C_{\text{in}}$ \\
\bottomrule
\end{tabular}}
\end{table}

\noindent
Each choice is both \emph{mixing-aware} (it performs exactly the mixing its role requires and nothing more) and \emph{compute-aware} (it is the cheapest operator for that role). By contrast, the depthwise-separable $Q/K/V$ used by CvT~\cite{wu_cvt_2021} adds an $HW\,C^2$ pointwise term of channel mixing before an attention that will mix channels anyway, which the ablation in Section~\ref{subsec:ablation} shows is both slower and less accurate. Because channel attention is global at linear spatial cost, no windowing, local-interaction modules, or gating are needed to control complexity.

\subsection{ADViT Architecture}
\label{subsec:arch}

\begin{figure}[t]
    \centering
    \includegraphics[width=\linewidth]{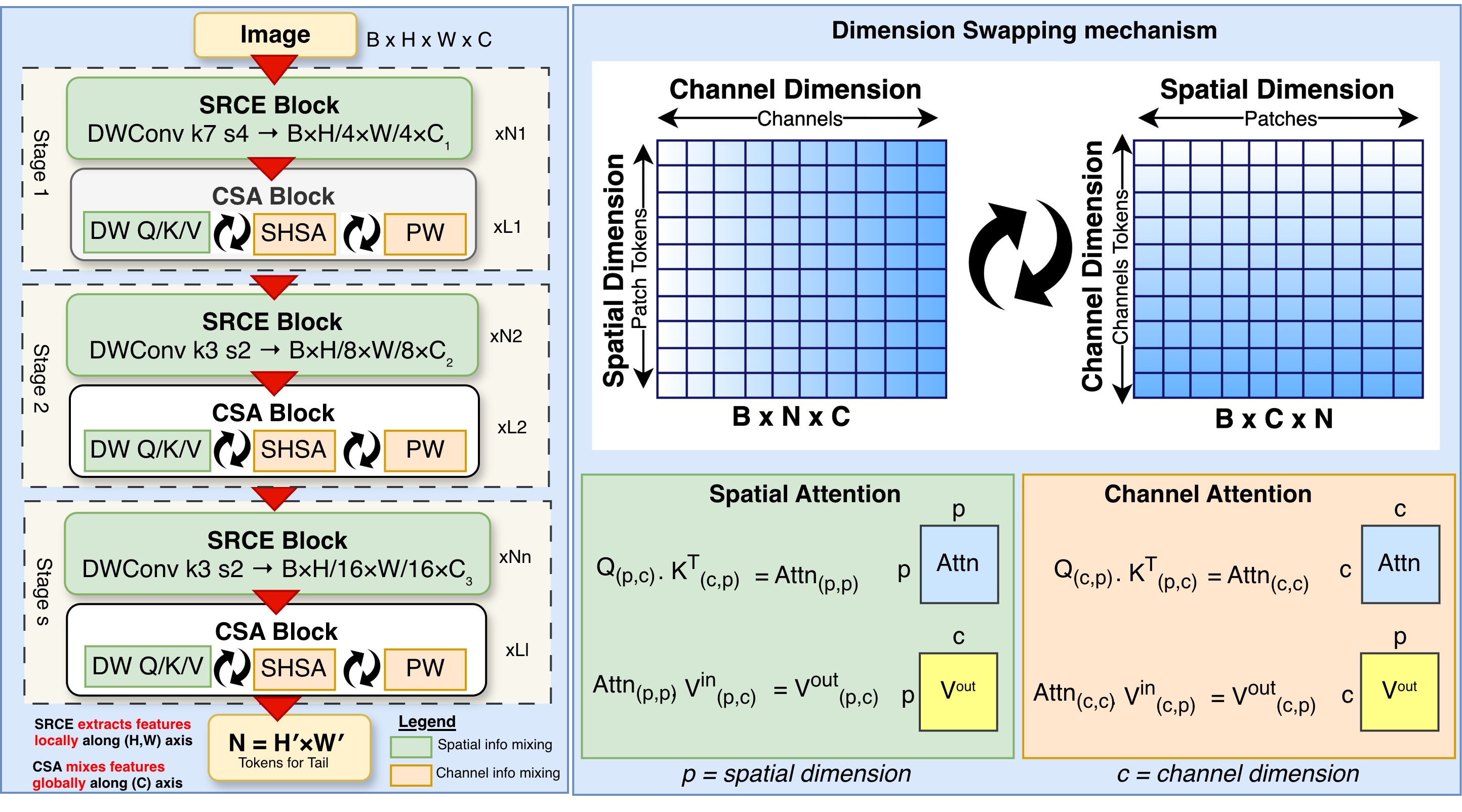}
    \caption{\textbf{ADViT schedule and dimension-swap mechanism.}
    Channel-token attention transposes the token and channel axes so that attention is computed over a $C{\times}C$ matrix rather than an $N{\times}N$ one.
    ADViT applies channel attention at high resolution and spatial MHSA at the lowest resolution, where the two mixers are complementary.}
    \label{fig:arch}
\end{figure}

ADViT is a three-stage hierarchical backbone (Fig.~\ref{fig:arch}) built from SRCE blocks for spatial extraction and CSA blocks for channel self-attention, with spatial MHSA reserved for the last, low-resolution stage.

\paragraph{SRCE block for spatial extraction.}
Each stage begins with one or more SRCE blocks performing pure spatial extraction by depthwise convolution,
\begin{equation}
\tilde{X} = \mathrm{Norm}\!\bigl(\phi\bigl(\mathrm{DW}_{k,s}(X)\bigr)\bigr),
\label{eq:srce}
\end{equation}
where $\mathrm{DW}_{k,s}$ is a depthwise convolution with kernel $k$ and stride $s$, and $\phi$ is GELU. When $s{>}1$, SRCE also downsamples. SRCE contains \emph{no} pointwise projection, so extraction and channel mixing are fully decoupled.

\paragraph{CSA block for channel self-attention.}
CSA is the primary representation learner. It forms $Q,K,V$ with depthwise-only operators and applies single-head self-attention over channels,
\begin{equation}
Q,K,V{=}\mathrm{Mat}(\mathrm{DW}_k(X))\!\in\!\mathbb{R}^{C\times HW},
\label{eq:qkv}
\end{equation}
where $\mathrm{Mat}$ flattens the spatial axes so each channel is a length-$HW$ token ($d{=}HW$), and $Q,K$ are normalized with a learnable scale, as in cross-covariance attention. The attention weights $A\!\in\!\mathbb{R}^{C\times C}$ of Eq.~\eqref{eq:channel_attn} are broadcast over all positions, $Y_{h,w,:}{=}A\,V_{h,w,:}$, and the block uses a pre-norm residual with a pointwise MLP,
\begin{equation}
X'{=}\mathrm{LN}(X{+}Y),\quad Z{=}\mathrm{LN}\!\bigl(X'{+}\mathrm{MLP}(X')\bigr).
\label{eq:block}
\end{equation}
A single head preserves holistic inter-feature interaction; depthwise-only $Q/K/V$ inject spatial context without the redundant channel mixing a pointwise projection would add.

\paragraph{Where spatial attention belongs.}
Channel and spatial attention are complementary, and the axis-decoupling lens is what makes their schedule visible. While $HW{\gg}C$, content-adaptive mixing is cheaper and more natural over channels; once $HW$ is small (\eg{} $14{\times}14$), a short stack of spatial MHSA blocks can add explicit spatial aggregation at modest cost. Coupled designs cannot defer that operator, because they rely on spatial mixing at every stage. ADViT's channel backbone already provides global interaction, so spatial MHSA waits until resolution is low.

\begin{table}[t]
\centering
\caption{\textbf{ADViT model family.} All variants use 3 stages, SRCE $k/s/p{=}[7,3,3]/[4,2,2]/[2,1,1]$, and depthwise-only $Q/K/V$ in CSA. The tail uses pointwise $Q/K/V$ with multi-head spatial attention.}
\label{tab:family}
\setlength{\tabcolsep}{5pt}
\renewcommand{\arraystretch}{1.15}
\footnotesize
\resizebox{\linewidth}{!}{%
\begin{tabular}{@{}lccccc@{}}
\toprule
\textbf{Model} & \textbf{Channels} & \textbf{CSA depth} & \textbf{MLP $r$} & \textbf{Tail} & \textbf{Params/FLOPs} \\
\midrule
ADViT-XXS & $[48,96,192]$  & $[1,2,4]$  & $[1,1,1]$ & 1 & 1.5\,M / 0.4\,G \\
ADViT-XS  & $[64,128,256]$ & $[1,2,5]$  & $[2,2,2]$ & 4 & 5.0\,M / 1.1\,G \\
ADViT-S   & $[64,192,384]$ & $[1,2,10]$ & $[4,4,4]$ & 5 & 16.8\,M / 4.2\,G \\
\bottomrule
\end{tabular}}
\end{table}

\paragraph{Model family.}
Table~\ref{tab:family} defines three scales sharing one topology. The cost follows directly from Eq.~\eqref{eq:projection}: SRCE is $O(HWCk^2)$ compute and $O(Ck^2)$ parameters, CSA is $O(HW\,C^2)$ per block with no $(HW)^2$ term, and the tail is $O((HW)^2C)$ but only at small $HW$.

\section{Experiments}
\label{sec:experiments}

We evaluate ADViT in the parameter- and data-limited regime, on ImageNet-1K, transfer, ADE20K segmentation, and controlled ablations.

\paragraph{Protocols.}
We report Top-1 accuracy (\%) for classification and mIoU for segmentation, with parameters (M) and FLOPs (G). ImageNet-1K~\cite{deng2009imagenet} follows the CvT recipe~\cite{wu_cvt_2021}; CIFAR-10/100~\cite{cifar10} and Tiny-ImageNet~\cite{le2015tiny} follow~\cite{zhang_depth-wise_2025} at $224^2$; transfer follows~\cite{guan_dynamic_2025}; ADE20K~\cite{zhou2017ade20k} uses UperNet~\cite{ferrari_unified_2018} via \texttt{mmsegmentation}~\cite{mmsegmentation_contributors_openmmlab_2020}. Hyperparameters are in the supplementary.
\rev{Every from-scratch result in Table~\ref{tab:fromscratch} and every ablation in Tables~\ref{tab:ablation}--\ref{tab:tailfree} is produced \emph{by us} under one identical recipe, so within-table comparisons are matched. Baseline numbers in Tables~\ref{tab:imagenet},~\ref{tab:transfer}, and~\ref{tab:ade20k} are quoted from the original papers unless stated otherwise; ADViT entries there are trained by us. Multi-seed results report mean\,$\pm$\,std over three seeds. The from-scratch and ImageNet runs inherit published spatial-ViT recipes without retuning, so a channel-primary backbone would likely benefit from further pipeline tuning; we leave that study to future work in order to isolate the architecture.}

\subsection{The Parameter- and Data-Limited Regime}
\label{subsec:scratch}

\begin{table}[t]
\centering
\caption{\textbf{From-scratch recognition at $224^2$.}
\best{Bold}\,=\,best, \secbest{underline}\,=\,second best per group.
P: parameters (M); F: FLOPs (G); C10/C100: CIFAR; TIN: Tiny-ImageNet.}
\label{tab:fromscratch}
\footnotesize
\setlength{\tabcolsep}{5pt}
\renewcommand{\arraystretch}{1.08}
\begin{tabular}{@{}lccccc@{}}
\toprule
\textbf{Model} & \textbf{P} & \textbf{F} & \textbf{C10} & \textbf{C100} & \textbf{TIN} \\
\midrule
\multicolumn{6}{@{}l@{}}{\gph{\textit{$>$10\,M params}}} \\
CvT-13            & 20.1 & 4.50 & \secbest{96.70} & \secbest{83.36} & \secbest{70.65} \\
ViT-S             & 21.7 & 4.61 & 95.09 & 73.97 & 61.23 \\
Swin-T            & 27.5 & 4.51 & 93.59 & 78.75 & 68.24 \\
\rowcolor{advitgray}
\textbf{ADViT-S}  & 16.8 & 4.20 & \best{97.20} & \best{83.59} & \best{72.28} \\[2pt]
\multicolumn{6}{@{}l@{}}{\gph{\textit{$\leq$10\,M params}}} \\
CaiT-XXS-12       &  6.4 & 1.30 & 92.02 & 73.43 & 59.17 \\
ViT-Ti            &  5.5 & 1.26 & \secbest{94.01} & 73.68 & 61.76 \\
\rowcolor{advitgray}
\textbf{ADViT-XS} & 5.0  & 1.18 & \best{96.10} & \best{79.61} & \best{66.34} \\
\rowcolor{advitgray}
\textbf{ADViT-XXS}& 1.5  & 0.40 & 93.71 & \secbest{73.96} & \secbest{63.16} \\
\bottomrule
\end{tabular}
\end{table}

Table~\ref{tab:fromscratch} reports from-scratch training, where no large-scale pretraining can compensate for a missing prior. ADViT-S leads every dataset despite being the smallest model in its group (16.8\,M \vs{} 20--28\,M), improving over CvT-13 by $1.6$ points on Tiny-ImageNet. In the sub-10\,M regime the gap widens: ADViT-XS beats ViT-Ti by $+2.1$ (C10), $+5.9$ (C100), and $+4.6$ (TIN), and ADViT-XXS at $1.5$\,M still matches or exceeds ViT-Ti on CIFAR-100 and Tiny-ImageNet. The advantage grows as the dataset gets harder and the budget shrinks, and the supplementary validation curves show the same ranking dynamically: ADViT climbs faster and plateaus higher on every dataset. \rev{Run-to-run variance is small here: over the 3-seed runs of Table~\ref{tab:tailfree} the largest std is $0.38$ (CIFAR-100) and $0.40$ (Tiny-ImageNet). As an indication rather than a significance test (those runs use the controlled rebuild, not the variants of Table~\ref{tab:fromscratch}), the margins above are $12$--$16\times$ that spread and the Table~\ref{tab:ablation} deltas ($-0.8$ to $-2.3$ on Tiny-ImageNet) are $2$--$6\times$.}

\subsection{ImageNet-1K Classification}
\label{subsec:imagenet}

\begin{table}[t]
\centering
\caption{\textbf{ImageNet-1K classification} at $224^2$, grouped by mixing strategy.
\rev{We do not claim state of the art: EdgeNeXt-B is more accurate at $18.5$\,M and $3.8$\,G. ADViT-S is competitive with models $20$--$75\%$ larger without windowing, dual-axis alternation, or gating.}}
\label{tab:imagenet}
\setlength{\tabcolsep}{6pt}
\renewcommand{\arraystretch}{1.06}
\footnotesize
\resizebox{\linewidth}{!}{%
\begin{tabular}{@{}lccc@{}}
\toprule
\textbf{Model} & \textbf{Params (M)} & \textbf{FLOPs (G)} & \textbf{Top-1 (\%)} \\
\midrule
\multicolumn{4}{@{}l@{}}{\gph{\textit{Spatial-attention}}} \\
DeiT-S~\cite{touvron_training_2021}        & 22.0 & 4.6  & 79.9 \\
Swin-T~\cite{liu_swin_2021}                & 29.0 & 4.5  & 81.3 \\
CvT-13~\cite{wu_cvt_2021}                   & 20.1 & 4.5  & 81.6 \\
PVTv2-B2~\cite{wang_pvtv2_2022}            & 25.4 & 4.0  & 82.0 \\
ConvNeXt-T~\cite{liu_convnet_2022}         & 29.0 & 4.5  & 82.1 \\
CSwin-T~\cite{dong_cswin_2022}             & 23.2 & 4.3  & 82.7 \\
UniFormer-S~\cite{li_uniformer_2023}       & 22.0 & 3.6  & 82.9 \\
\midrule
\multicolumn{4}{@{}l@{}}{\gph{\textit{Efficient ViTs}}} \\
MobileViT-XXS~\cite{mehta_mobilevit_2022}  &  1.3 & 0.4  & 70.4 \\
EdgeViT-XS~\cite{pan_edgevits_2022}        &  6.7 & 1.1  & 77.5 \\
EdgeViT-S~\cite{pan_edgevits_2022}         & 11.1 & 1.9  & 81.0 \\
\midrule
\multicolumn{4}{@{}l@{}}{\gph{\textit{Channel-aware / dual-axis}}} \\
SCAT-XS~\cite{zheng_lightweight_2023}      &  8.7 & 1.2  & 79.2 \\
SCHEME-12-S12~\cite{sridhar_scheme_2024}   & 21.0 & 3.4  & 82.0 \\
DaViT-T~\cite{ding_davit_2022}             & 28.3 & 4.5  & 82.8 \\
DCT-ViT-S~\cite{guan_dynamic_2025}         & 28.4 & 4.5  & 82.9 \\
\rev{EdgeNeXt-S~\cite{maaz_edgenext_2022}} & \rev{5.6}  & \rev{1.3} & \rev{79.4} \\
\rev{EdgeNeXt-B~\cite{maaz_edgenext_2022}} & \rev{18.5} & \rev{3.8} & \rev{82.5} \\
\midrule
\rowcolor{advitgray}
\textbf{ADViT-S (ours)} & \textbf{16.8} & 4.2 & 82.1 \\
\bottomrule
\end{tabular}}
\end{table}

At $16.8$\,M parameters and $4.2$\,G FLOPs, ADViT-S reaches $82.1\%$ Top-1 (Table~\ref{tab:imagenet}). It is competitive with spatial-attention models of similar size (Swin-T $81.3$, CvT-13 $81.6$, ConvNeXt-T $82.1$) and matches the strongest channel-aware models (DaViT-T $82.8$, DCT-ViT-S $82.9$, SCHEME-12-S12 $82.0$, \rev{EdgeNeXt-B $82.5$}) to within $0.8$ points. DaViT-T and DCT-ViT-S reach those accuracies with dual-axis alternation and dynamic gating at $\sim$$28$\,M parameters; ADViT-S stays within $0.8$ points at $20$--$40\%$ fewer parameters, without that machinery.
\rev{We do not claim state of the art. EdgeNeXt-B in particular is more accurate at fewer FLOPs, though it is a fused conv-transformer hybrid rather than an axis-decoupled backbone (Sec.~\ref{sec:related}). Our budget allowed one full ImageNet schedule, spent on ADViT-S, so the sub-$10$\,M comparison against EdgeNeXt-S is left open; the small-budget evidence is the matched-recipe regime of Table~\ref{tab:fromscratch}. The ImageNet claim is parity at fewer parameters, without additional modules, not superiority.}

\subsection{Transfer Learning}
\label{subsec:transfer}

\begin{table}[t]
\centering
\footnotesize
\setlength{\tabcolsep}{4pt}
\renewcommand{\arraystretch}{1.1}
\caption{\textbf{Transfer learning} (Top-1 \%) on Oxford Flowers-102 and Oxford-IIIT Pets with ImageNet-1K pretrained backbones, following~\cite{guan_dynamic_2025}.
$\uparrow$\,=\,fine-tuned at higher resolution. \best{Bold}\,=\,best, \secbest{underline}\,=\,second best.}
\label{tab:transfer}
\begin{tabular}{@{}lcccc@{}}
\toprule
\textbf{Model} & \textbf{Params (M)} & \textbf{FLOPs (G)} & \textbf{Flowers} & \textbf{Pets} \\
\midrule
ResNet-50              & 25.6 &  4.1 & 98.2 & {--} \\
EfficientNet-B5        & 28.0 & 10.3 & 98.5 & {--} \\
ViT-B/16               & 86.6 & 18.7 & 89.5 & 93.8 \\
DeiT-B                 & 86.6 & 17.3 & 98.4 & {--} \\
CeiT-S                 & 24.2 &  5.1 & 98.2 & \secbest{94.6} \\
DCT-ViT-S$^{\uparrow 384}$ & 28.4 & 12.7 & \secbest{98.6} & 94.5 \\
\rowcolor{advitgray}
\textbf{ADViT-S (ours)} & 16.8 & 4.2 & \best{98.7} & \best{94.8} \\
\bottomrule
\end{tabular}
\end{table}

Fine-tuning the same ImageNet-1K pretrained backbones (Table~\ref{tab:transfer}), ADViT-S is best on Pets ($94.8$) and Flowers ($98.7$) despite being the smallest model ($2$--$5\times$ fewer parameters), and it surpasses DCT-ViT-S even when the latter is fine-tuned at $384^2$.

\subsection{Semantic Segmentation on ADE20K}
\label{subsec:seg}

\begin{table}[t]
\centering
\caption{\textbf{ADE20K segmentation} with UperNet~\cite{ferrari_unified_2018}.
MS\,=\,multi-scale. FLOPs at $512{\times}2048$; FPS on a single GPU.}
\label{tab:ade20k}
\setlength{\tabcolsep}{4pt}
\renewcommand{\arraystretch}{1.08}
\footnotesize
\resizebox{\linewidth}{!}{%
\begin{tabular}{@{}lccccc@{}}
\toprule
\textbf{Backbone} & \textbf{mIoU} & \textbf{MS} & \textbf{Params} & \textbf{FLOPs} & \textbf{FPS} \\
\midrule
Swin-T~\cite{liu_swin_2021}            & 39.89 & 40.80 & 59.94 & 945.66  & \best{92.7} \\
DeiT-S~\cite{touvron_training_2021}    & 41.88 & \best{42.84} & 57.99 & 1216.82 & 80.4 \\
DaViT-T~\cite{ding_davit_2022}         & 41.19 & 42.13 & 60.01 & 942.05  & 84.7 \\
XCiT-S12/16~\cite{ali_xcit_2021}       & 41.37 & 41.96 & 61.88 & 1071.82 & 84.2 \\
\rowcolor{advitgray}
\textbf{ADViT-S}                       & \best{42.00} & \secbest{42.55} & \best{46.15} & \best{814.24} & \secbest{91.9} \\
\bottomrule
\end{tabular}}
\end{table}

\begin{figure}[t]
  \centering
  \includegraphics[width=\linewidth]{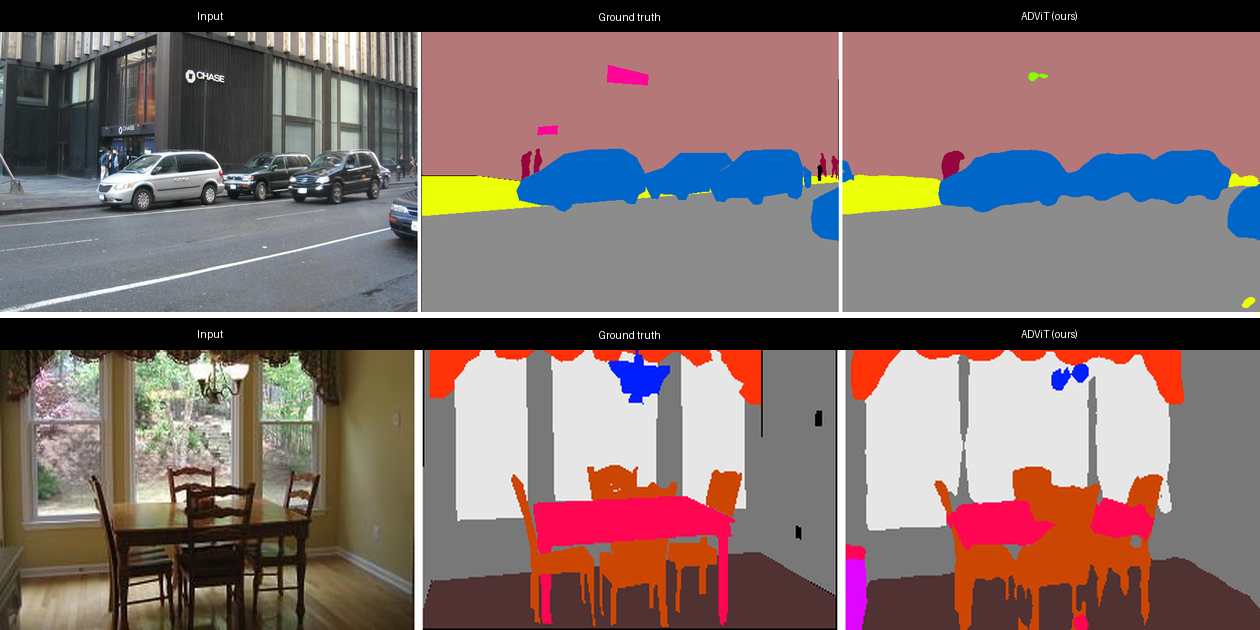}
  \caption{\rev{\textbf{Qualitative ADE20K segmentation.} Input, ground truth, and ADViT-S (UperNet) on two validation scenes. Channel-primary mixing recovers correct semantic layout across indoor and outdoor scenes without windowed or dual-axis spatial-attention machinery. Failure modes are also visible: small and thin structures (distant pedestrians, signage, chair legs) are merged or missed, consistent with a 42.0 mIoU operating point.}}
  \label{fig:seg}
\end{figure}

\begin{figure}[t]
  \centering
  \includegraphics[width=\linewidth]{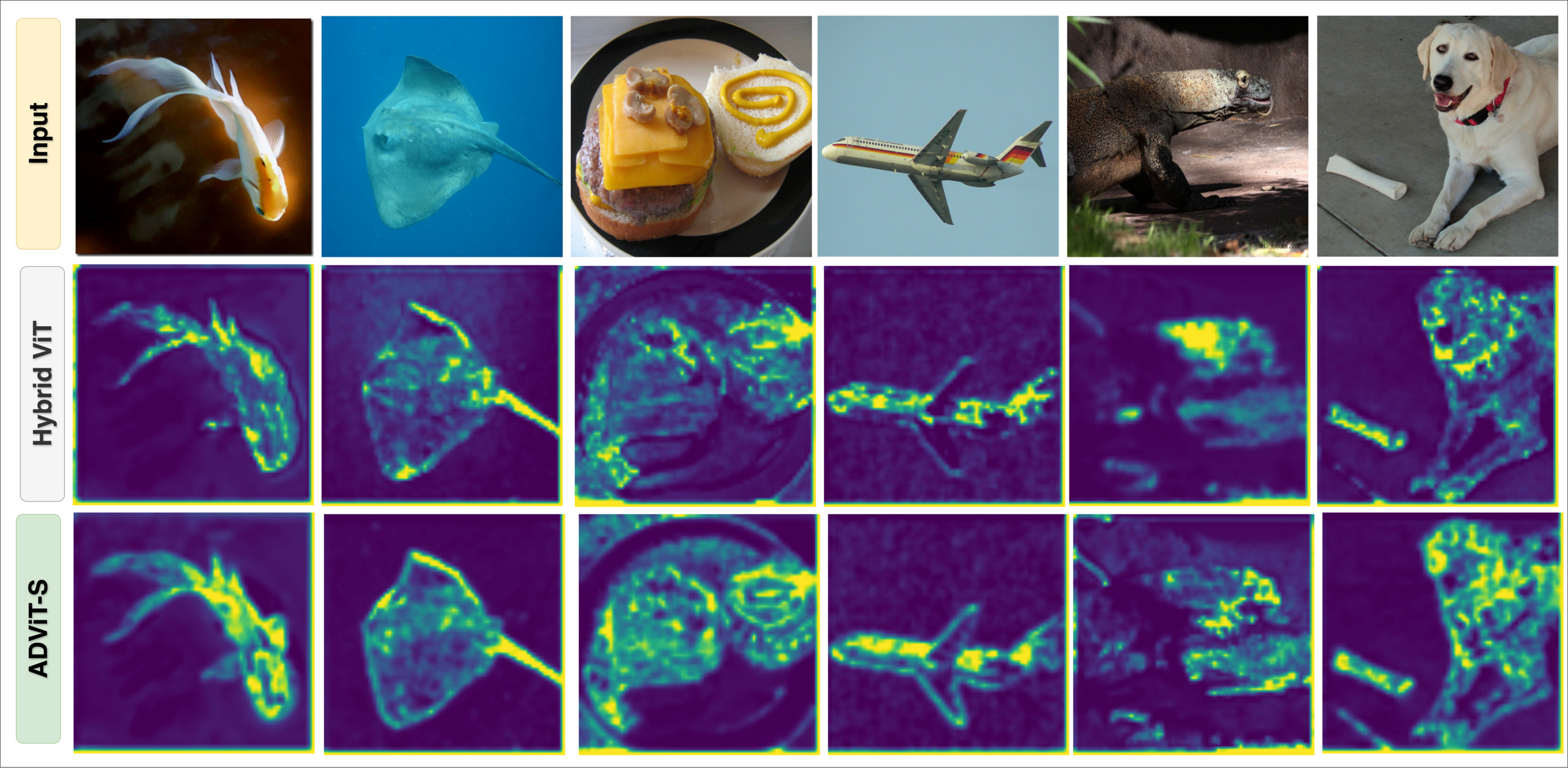}
  \caption{\textbf{Attention visualization} (DINO-style rollout at the final $14{\times}14$ stage, bilinearly upsampled). ADViT-S produces concentrated, object-focused attention with less background diffusion than a comparable spatial-attention hybrid ViT.}
  \label{fig:attn}
\end{figure}

ADViT-S obtains the best single-scale mIoU ($42.00$) and the best accuracy-efficiency trade-off (Table~\ref{tab:ade20k}): $814$\,G FLOPs against $942$--$1217$\,G, $20\%$ fewer parameters, and $91.9$ FPS, on par with Swin-T and faster than DaViT-T and XCiT-S12/16. Staged depthwise extraction keeps the spatial grid intact; channel attention supplies global context at linear spatial cost. Figure~\ref{fig:seg} shows the resulting predictions on indoor and outdoor scenes. Figure~\ref{fig:attn} compares attention rollouts against CvT-13, the spatial-attention hybrid closest to ADViT-S in capacity ($\sim$$20$\,M), training recipe, and staged conv-plus-attention design; matching these factors isolates the mixing axis. ADViT-S yields sharper, more object-focused activations across categories, while the hybrid baseline is more diffuse, consistent with channel-primary mixing concentrating capacity on semantically relevant features.
 \rev{Throughput, however, is resolution-dependent and we scope the claim accordingly: depthwise convolutions are cheap in FLOPs but memory-bound, so savings convert into speed only where the quadratic spatial term dominates. At $224^2$ it does not, and the parameter and FLOP savings there do not translate into a wall-clock advantage. We claim parameter, FLOP, and high-resolution-throughput efficiency, not low-resolution latency.}

\subsection{Ablations and the Tail-Free Backbone}
\label{subsec:ablation}

\begin{table}[t]
\centering
\caption{\textbf{Ablations on Tiny-ImageNet} (from scratch, $224^2$, ADViT-XS scale). Each row changes one design choice relative to the full model.}
\label{tab:ablation}
\footnotesize
\setlength{\tabcolsep}{4pt}
\renewcommand{\arraystretch}{1.06}
\begin{tabular}{@{}lcccccc@{}}
\toprule
\textbf{Variant} & \textbf{SRCE} & \textbf{CSA} & \textbf{Tail} & \textbf{P} & \textbf{F} & \textbf{Top-1} \\
\midrule
ViT-Ti (baseline)        & --          & --          & --          & 5.5 & 1.26 & \rev{61.76} \\
\midrule
\rowcolor{advitgray}
ADViT-XS (full)          & \checkmark & \checkmark & \checkmark & 5.1 & 1.18 & \best{66.34} \\
\;\;w/o SA tail          & \checkmark & \checkmark & --          & 1.9 & 0.70 & 65.36 \\
\;\;w/o channel attn     & \checkmark & --          & \checkmark & 5.1 & 2.90 & 64.00 \\
\;\;DW-sep $Q/K/V$       & \checkmark & \checkmark & \checkmark & 6.1 & 1.70 & 65.50 \\
\;\;spatial-first sched. & \checkmark & \checkmark & \checkmark & 5.1 & 2.90 & 65.00 \\
\bottomrule
\end{tabular}
\end{table}

Table~\ref{tab:ablation} isolates each design choice. Replacing channel attention with spatial attention causes the largest drop ($-2.3$) \emph{and} inflates FLOPs from $1.18$ to $2.90$\,G; \rev{SRCE and the tail are held fixed in that row, so it isolates channel-primary mixing from the convolutional inductive bias the two variants share.} Swapping depthwise-only $Q/K/V$ for depthwise-separable projections costs $-0.8$ at extra parameters, the redundant pre-attention channel mixing the analysis warns against. Reversing the schedule to spatial-first degrades by $-1.3$ at higher FLOPs. Removing the tail still yields $65.36\%$ at $1.9$\,M parameters.

\begin{figure}[t]
  \centering
  \includegraphics[width=0.74\linewidth]{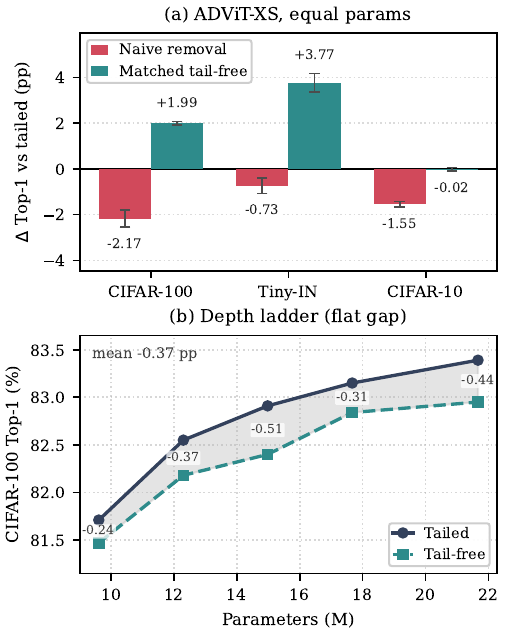}
  \caption{\rev{\textbf{Channel-primary mixing on its own.}
  \textbf{(a)}~ADViT-XS, equal params: matched tail-free exceeds tailed (${+}1.99$ CIFAR-100, ${+}3.77$ Tiny-ImageNet; 3 seeds); naive removal always loses.
  \textbf{(b)}~CIFAR-100 depth ladder (fixed width/MLP): small, \emph{flat} gap (mean ${-}0.37$\,pp) that does not grow with capacity.}}
  \label{fig:tailfree_fig}
\end{figure}

\rev{To test whether channel-primary mixing is competitive, we compare, under one recipe, the tailed model, \emph{naive} deletion of the spatial tail, and a \emph{parameter-matched} tail-free model that reallocates the tail budget into stage-3 channel depth (Table~\ref{tab:tailfree}, Fig.~\ref{fig:tailfree_fig}). Absolute numbers differ slightly from Tables~\ref{tab:fromscratch}--\ref{tab:imagenet} because this study uses one controlled rebuild; comparisons here are matched. The tail is $3.2$\,M of $5.1$\,M ($62\%$) at XS but $5.2$\,M of $16.8$\,M ($31\%$) at S. Naive removal always loses; reinvesting the budget is the right test. At XS, that reinvestment \emph{exceeds} the tailed model by $+1.99{\pm}0.08$ on CIFAR-100 and $+3.77{\pm}0.40$ on Tiny-ImageNet, well outside seed variance. CIFAR-10 is already saturated near $96\%$ and stays a tie. At S, where there is less tail budget to reallocate, matched tail-free is within $-0.33$ on CIFAR-10, $-0.55$ on CIFAR-100, and $-0.84$ on Tiny-ImageNet. On ImageNet-1K under a \emph{100-epoch} matched schedule it reaches $79.06\%$ vs.\ tailed $79.12\%$ ($-0.06$), whereas naive removal drops to $77.94\%$ ($-1.18$). A CIFAR-100 depth ladder at fixed width and MLP ratio shows a small, \emph{flat} gap (mean $-0.37$\,pp) that does not grow with capacity; width is the wrong scaling axis here, because a $C{\times}C$ mixer grows quadratically in $C$. Together these support that channel-primary mixing is competitive on its own, with last-stage spatial attention complementary rather than required.}
\begin{table}[t]
\centering
\caption{\rev{\textbf{Tail-free ADViT} from scratch (mean\,$\pm$\,std over 3 seeds).
\emph{Naive} deletes the spatial tail (4 blocks at XS, 5 at S, per Table~\ref{tab:family}); \emph{matched} reallocates that budget into stage-3 channel depth at equal parameters.
The tail is $3.2$\,M of $5.1$\,M ($62\%$) at XS and $5.2$\,M of $16.8$\,M ($31\%$) at S.
\best{Bold}\,=\,best, \secbest{underline}\,=\,second best per scale.}}
\label{tab:tailfree}
\scriptsize
\setlength{\tabcolsep}{4.5pt}
\renewcommand{\arraystretch}{1.12}
\begin{tabular}{@{}lccc@{}}
\toprule
\textbf{Configuration} & \textbf{C10} & \textbf{C100} & \textbf{TIN} \\
\midrule
\multicolumn{4}{@{}l@{}}{\gph{\textit{ADViT-XS scale}}} \\
Tailed (full)     & $\mathbf{96.02{\pm}0.07}$ & \underline{$79.32{\pm}0.07$} & \underline{$66.19{\pm}0.32$} \\
Naive w/o tail    & $94.47{\pm}0.13$ & $77.15{\pm}0.38$ & $65.46{\pm}0.34$ \\
\rowcolor{advitgray}
Matched tail-free & \underline{$96.00{\pm}0.08$} & $\mathbf{81.31{\pm}0.08}$ & $\mathbf{69.96{\pm}0.40}$ \\
\midrule
\multicolumn{4}{@{}l@{}}{\gph{\textit{ADViT-S scale}}} \\
Tailed (full)     & $\mathbf{96.86{\pm}0.10}$ & $\mathbf{83.33{\pm}0.05}$ & $\mathbf{73.05{\pm}0.11}$ \\
Naive w/o tail    & $96.14{\pm}0.16$ & $82.53{\pm}0.10$ & $72.06{\pm}0.13$ \\
\rowcolor{advitgray}
Matched tail-free & \underline{$96.53{\pm}0.14$} & \underline{$82.78{\pm}0.20$} & \underline{$72.21{\pm}0.14$} \\
\bottomrule
\end{tabular}
\end{table}

\paragraph{When does channel-first help?}
The experiments trace a consistent boundary: channel-primary mixing helps most where spatial attention is weakest, under small parameter budgets, where the margin over ViT-Ti and CvT-13 widens as capacity shrinks (Table~\ref{tab:fromscratch}); under limited labels, where the from-scratch advantage grows from CIFAR-10 to Tiny-ImageNet; and at high resolution, where linear-cost mixing yields the best segmentation trade-off (Table~\ref{tab:ade20k}). The tail-free study sits on the same boundary: reallocating the tail into channel depth helps most at XS and remains competitive at S (Table~\ref{tab:tailfree}). On ImageNet-1K the result is parity rather than dominance (Sec.~\ref{subsec:imagenet}).

\section{Conclusion}
\label{sec:conclusion}

We asked how far a deliberately simple idea can go: making content-adaptive mixing channel-primary and \rev{decoupling} it from spatial extraction, rather than interleaving it into a spatial-first pipeline. ADViT is simpler by construction, without the auxiliary modules of prior channel-aware designs, yet remains competitive and is most advantageous when parameters, compute, or data are limited. The axis-decoupling lens also gives a vocabulary for where adaptive mixing should live. Next steps are scaling beyond ADViT-S, detection, and mixer-specific training recipes.

\clearpage
\section*{Acknowledgments}
This publication has emanated from research conducted with the financial support of Taighde \'Eireann -- Research Ireland under Grant number 18/CRT/6183.

{
    \small
    \bibliographystyle{ieeenat_fullname}
    \bibliography{main}
}

\end{document}